\documentclass{article} % For LaTeX2e
\usepackage{iclr2027_conference,times}

\usepackage{amsmath,amsfonts,bm}

\def\eqref#1{equation~\ref{#1}}
\def\1{\bm{1}}

\DeclareMathAlphabet{\mathsfit}{\encodingdefault}{\sfdefault}{m}{sl}
\SetMathAlphabet{\mathsfit}{bold}{\encodingdefault}{\sfdefault}{bx}{n}

\usepackage{hyperref}
\usepackage{url}
\usepackage{graphicx}
\usepackage{enumitem}
\usepackage{amsthm}
\usepackage{amsmath}
\usepackage{amssymb}
\newtheorem{theorem}{Theorem}
\usepackage{tcolorbox}
\usepackage{booktabs}
\usepackage{multirow}
\usepackage{wrapfig}
\usepackage{subcaption}
\usepackage{algorithm}
\usepackage{algpseudocode}
\usepackage[dvipsnames]{xcolor}
\usepackage[table]{xcolor}
\usepackage{tabularx}
\usepackage{makecell}

\definecolor{myblue}{HTML}{6C8EBF}
\definecolor{gblue}{HTML}{7EA6E0}
\definecolor{gred}{HTML}{EA6B66}

\title{
Get Hint-Augmented GRPO Done Right\\Through Activation Steering
}

\iclrfinalcopy
\author{Khiem Le, Ting Hua, Nitesh V. Chawla\\
University of Notre Dame, IN, USA
}

\begin{document}

\maketitle

\begin{abstract}
GRPO has become the dominant reinforcement learning method for eliciting reasoning capability in LLMs using verifiable rewards. However, it suffers from a fundamental failure mode: when a training question exceeds the base model's current capability, all generated responses fail and receive zero reward, yielding no gradient signal and stalling further capability expansion. An emerging class of GRPO extensions, termed Hint-Augmented GRPO (HA-GRPO), adds a “hint” to the context to reduce the difficulty of the question, thereby producing a usable gradient signal. Nevertheless, existing HA-GRPO methods share a key issue: no hint is available at inference time to append to the question, creating a training-inference context mismatch that leads to inconsistent performance gains. In this work, we reinterpret the function of hinting – not as reducing question difficulty, but as steering model behavior – and propose ASH (Activation Steering for Hinting), a novel HA-GRPO method built on this view. During training, the model is presented with both hinted and plain questions, while a trainable steering vector is optimized to widen the margins between the model's generation probabilities for responses to the hinted version relative to the plain version, thereby capturing the question-agnostic component of the behavioral shift induced by the hint. During inference, the trained steering vector is directly injected into the model's activation, reproducing the hint-conditioned behavior without requiring any hint in the context. Experiments with four LLMs (Qwen3-1.7B/4B-Base, Llama-3.2-1B/3B-Instruct) show that ASH consistently improves pass@$k$ on competition-level mathematics benchmarks (AMC, OlympiadBench, AIME24, AIME25) and out-of-distribution benchmarks (Minerva, GPQA-Diamond). Notably, it improves average pass@8 of Qwen3-1.7B-Base and Qwen3-4B-Base by \textbf{2.60}\% and \textbf{2.21}\%.
\end{abstract}

\section{Introduction}
Recently, reinforcement learning with verifiable rewards \citep{lambert2024tulu} has emerged as the key driver for eliciting reasoning capability in Large Language Models (LLMs), particularly in domains such as mathematics and coding, where correctness of final answers can be scored automatically \citep{jaech2024openai, guo2025deepseek}. Among various methods, Group Relative Policy Optimization (GRPO) \citep{shao2024deepseekmath} is widely adopted; it eliminates the critic model and estimates advantages for a group of generated responses through group-relative reward normalization. Although this design offers impressive computational efficiency, it fundamentally fails when a training question is \textbf{too difficult} for the base model: all generated responses fail and receive zero reward; their advantages consequently vanish, and the question yields no gradient signal, blocking any capability expansion beyond the base model \citep{yue2025does, akgul2026rethinking}. This limitation motivates the need for methods that can produce gradient signals on such too-difficult questions.

One simple approach is to allocate more generation budget to too-difficult questions, in hopes of eventually reaching rare successful responses \citep{shrivastava2026, yu2026dapo}; however, this multiplies computational cost. A second approach uses a proxy for successful responses, such as the model's confidence \citep{han2026self} or certainty \citep{le2026no}, but recovers only \underline{indirect} gradient signals. A third, more promising approach reduces the difficulty of the questions themselves by adding a “hint” to the context, giving rise to a class of methods we term Hint-Augmented GRPO (HA-GRPO) \citep{li2026questa, liao2026self, xia2026}. For example, given a training question, QUESTA \citep{li2026questa} uses the first 50\% of tokens from the solution as a hint and augments the question to guide the model toward successful responses. Follow-up HA-GRPO methods employ more subtle hint-extraction mechanisms; however, they all share a critical, unresolved issue: a training-inference context mismatch, since no hint is available at inference time to append to the question. Intuitively, these methods allow the model to \textit{cheat} by leveraging assistance during training, whereas no such assistance is permitted during inference (i.e., the real exams).

\begin{figure}[!t]\centering
\begin{subfigure}{0.49\textwidth}
\centering
\includegraphics[width=0.866\linewidth]{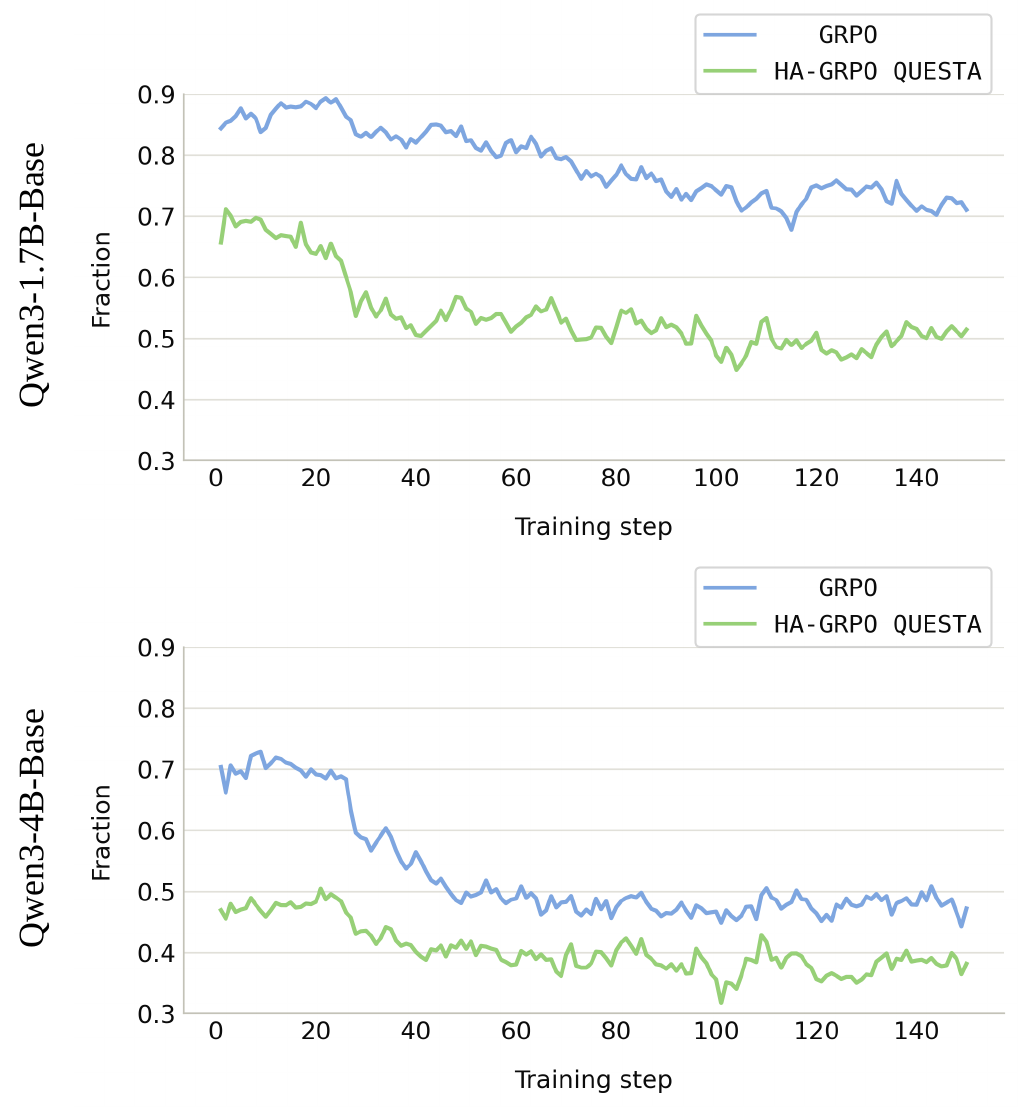}
\caption{The fraction of too-difficult questions per batch over the course of training.}\label{fig-1a}
\end{subfigure}
\hfill
\begin{subfigure}{0.49\textwidth}
\centering
\includegraphics[width=0.866\linewidth]{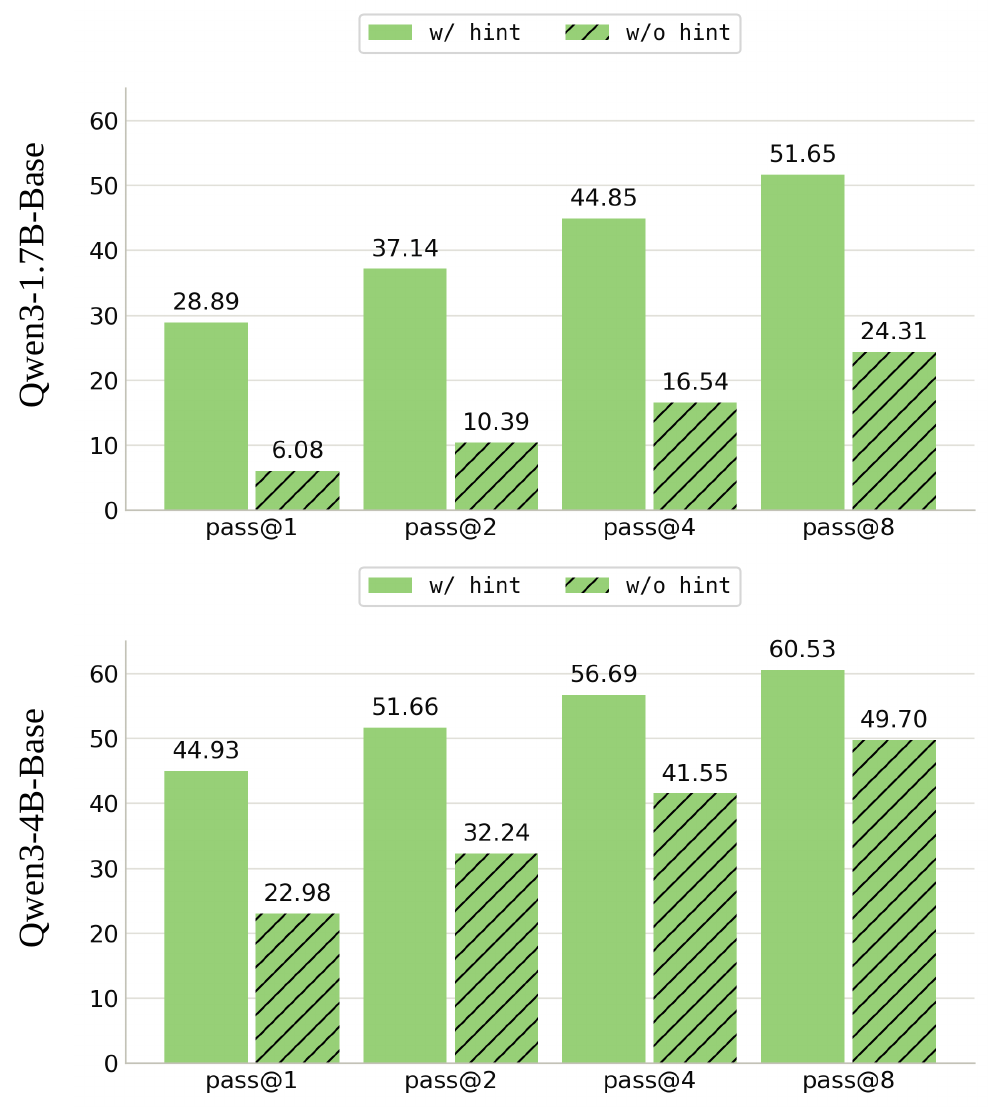}
\caption{pass@$k$ of QUESTA-trained models, with and without the hint in the questions.}\label{fig-1b}
\end{subfigure}
\caption{The impact of the training-inference context mismatch. (a) QUESTA substantially lessens the fraction of too-difficult questions per batch over the course of training, relative to GRPO. (b) pass@$k$ of QUESTA-trained models drops sharply without the hint at inference time.}\end{figure}

To characterize the impact of this training-inference context mismatch, we run preliminary experiments using GRPO and QUESTA to train Qwen3-1.7B-Base and Qwen3-4B-Base on a subset of the OpenR1-MATH-220k dataset \citep{openr1} (details in Section \ref{sec-exps}). As shown in Figure \ref{fig-1a}, QUESTA substantially lessens the fraction of too-difficult questions per batch over the course of training, relative to GRPO, aligned with expectations. We then evaluate QUESTA-trained models on training data, with and without the hint in the questions. As shown in Figure \ref{fig-1b}, pass@$k$ drops sharply without the hint at inference time; for example, pass@8 falls by \textbf{27.34}\% and \textbf{10.83}\% for Qwen3-1.7B-Base and Qwen3-4B-Base, respectively. These huge performance gaps indicate that models trained with QUESTA rely heavily on the hint as a crutch during training: much of the capability boost that hinting unlocked is lost, and the models consequently generalize poorly to the hint-free setting they will actually face during inference.

To mitigate this issue, we reframe the function of hinting: rather than viewing it as reducing question difficulty, as existing HA-GRPO methods do, we view it as steering model behavior. This view situates hinting within a broader paradigm of context-based control, alongside few-shot in-context learning \citep{todd2024}, which steers behavior through demonstrations, and instruction following \citep{stolfo2025improving}, which steers behavior through explicit directives. Under this lens, the hint in QUESTA and other methods steers model behavior toward a higher-quality reasoning space rather than reducing question difficulty, suggesting that hint-reliance stems not from the hint itself, but from how the information is introduced during training yet withdrawn during inference.

Building on this view, we propose ASH (Activation Steering for Hinting), which leverages the hint to steer model behavior toward a higher-quality reasoning space not through the context, but through the model's activation. During training, both hinted and plain questions are presented, and a trainable steering vector is added to the model's activation at a particular layer and optimized to widen the margins between the generation probabilities the model assigns to responses under the hinted version relative to the plain version. This drives the steering vector to capture, within the activation space, the question-agnostic component of the behavioral shift that the hint induces, which is precisely the component that transfers to unseen questions. During inference, the trained steering vector is directly injected into the model's activation, reproducing the hint-conditioned behavior without ever exposing the model to a hint. As the benefit of the hint is distilled into the steering vector that remains available at inference time, ASH eliminates the training-inference context mismatch, allowing the model to retain a substantial part of the capability boost hinting unlocked rather than losing it.

Our contributions are summarized as follows:
\begin{itemize}[topsep=0em, leftmargin=*, noitemsep]
\item \textbf{The training-inference context mismatch issue.} We identify a critical issue that existing HA-GRPO methods have overlooked and conduct preliminary experiments to characterize its impact.
\item \textbf{ASH.} Our novel HA-GRPO method mitigates the identified issue through the lens of steering model behavior. ASH internalizes the hint's behavioral shift in a trainable steering vector that remains available and can be used at inference time.
\item \textbf{Evaluation.} ASH consistently improves pass@$k$ across four LLMs. On average, pass@8 gains reach \textbf{1.97}\%/\textbf{2.25}\% on competition-level mathematics and \textbf{3.84}\%/\textbf{2.13}\% on out-of-distribution benchmarks for Qwen3-1.7B/4B-Base, and reach \textbf{1.86}\%/\textbf{2.15}\% on competition-level mathematics and \textbf{2.80}\%/\textbf{2.11}\% on out-of-distribution benchmarks for Llama-3.2-1B/3B-Instruct.
\item \textbf{In-depth analysis.} We showcase that the trained steering vector can transfer effectively to other reasoning domains such as logical question answering and coding.
\end{itemize}

\section{Related Work}
In this section, we review prior work on GRPO advances before narrowing our focus to the Hint-Augmented GRPO (HA-GRPO) class. Then, we briefly discuss activation steering research.

\textbf{GRPO Advances} - A line of work aims to make GRPO better. For instance, given a group of generated responses per training question, SAGE-GRPO \citep{huang2026does} seeks to control the lengths of these responses. Dr. GRPO \citep{liu2025understanding} aims to overcome the question-level difficulty bias in GRPO, whereas GRPO-LEAD \citep{zhang2025grpo} embraces this bias by reweighting each question by its difficulty, estimated via the inverse success rate of the responses. Despite rapid progress \citep{surana2026generate}, a fundamental failure mode of GRPO persists: when a training question is \textbf{too difficult} for the base model, all generated responses fail and receive zero reward; their advantages vanish, and the question then yields no gradient signal, blocking any capability gain beyond the base model \citep{yue2025does, akgul2026rethinking}. To address this limitation, SA-GRPO \citep{han2026self} and GRPO-ZVP \citep{le2026no} leverage the model's confidence and certainty as a proxy to complement the response-level rewards. However, the proxy is decoupled from correctness, so the gradient signal recovered is not guaranteed to point in a direction of enhancing capability.

\textbf{HA-GRPO} - A class of GRPO extensions attempts to reduce the difficulty of such questions by adding a “hint” to the context, based on their reference solution. For example, given a training question, QUESTA \citep{li2026questa} uses the first 50\% of tokens from the solution as a hint and augments the question to guide the model toward successful responses. BREAD \citep{zhang2026bread} uses the first 50\% of sentences, separated by “\verb|\n|”. Follow-up methods introduce more subtle mechanisms to extract the hint. Notably, \citet{zhang2026scaf, wang2026don} draw on the scaffolding theory to construct a hierarchical, progressive hint rather than a solution prefix, though this strategy requires a powerful, proprietary teacher model (e.g., GPT). To relax this dependence, SAGE \citep{liao2026self} extracts an adaptive hint from the model itself, conditioning not only on the solution but also on generated responses; HILL \citep{xia2026}, meanwhile, concurrently creates a duplicated copy of the model to serve as the hint provider. Although these two methods eliminate the dependence on a proprietary teacher model, both nearly double the computational cost. Importantly, existing HA-GRPO methods focus on hint-extraction mechanisms while overlooking the issue that no hint is available at inference time to append to the question. In this work, we mitigate this training-inference context mismatch through the lens of steering model behavior.

\textbf{Activation Steering} - An activation-based paradigm for controlling the behavior of LLMs \citep{zou2023representation, wehner2025taxonomy}. This is done by first searching for “steering vectors” within the activation space of particular layers in LLMs' architecture. During inference, these steering vectors are directly injected back to alter model generations. \citet{turner2023} explicitly created steering vectors associated with contrastive concepts, e.g., “Love” and “Hate”. Building on this idea, \citet{todd2024} construct steering vectors that encode demonstrations to enhance few-shot in-context learning, and \citet{stolfo2025improving} construct steering vectors that encode explicit directives to enhance instruction following. In common, such techniques construct steering vectors offline from pairs of contrastive prompts, limiting their effectiveness on complex reasoning tasks. In this work, we build on this paradigm and construct a steering vector for hinting that is made trainable online by design.

\section{Preliminaries}
We formalize GRPO and its failure on too-difficult questions, and introduce the Hint-Augmented GRPO (HA-GRPO) class built on top of it. Then, we describe how activation steering works.

\textbf{GRPO.} Recent efforts to strengthen the reasoning capability of LLMs have increasingly depended on GRPO, which estimates advantages from group-relative reward normalization. Let $\pi_\theta$ represent the model, which is parameterized by $\theta \in \mathbb{R}^d$. From the training data $\mathcal{D}$, sample the question $q$, the model generates $G$ responses to it as a group $\{o_i \sim \pi_\theta (\cdot|q)\}_{i=1, ..., G}$. Denote their rewards as $\{R_i\}_{i=1, ..., G}$, the advantages of $G$ responses are computed via:
\begin{equation*}
\{A_i = \frac{R_i - \text{mean}(\{R_i\}_{i=1, ..., G})}{\text{std}(\{R_i\}_{i=1, ..., G})+\varepsilon}\}_{i=1, ..., G}. 
\end{equation*}
Next, the model is updated by minimizing a loss (simplified) defined based on these advantages:
\begin{gather}
J(\theta; q) = - \left[ \frac{1}{G} \sum_{i=1}^{G} \frac{1}{|o_i|} \sum_{t=1}^{|o_i|} \left( r_\theta(o_{i, t})A_i \right) \right], \nonumber \\
\text{where} \quad r_\theta(o_{i, t}) = \frac{\pi_\theta(o_{i, t} | q, o_{i, < t})}{[\pi_\theta(o_{i, t} | q, o_{i, < t})]_{\text{not\_updated}}}. 
\label{eqn-1}
\end{gather}

Despite its strengths, this group-relative reward normalization design has a fragility that blocks capability expansion. When the question $q$ is too difficult (all $G$ generated responses fail), the reward satisfies $R_i = \text{mean}(\{R_i\}_{i=1, ..., G})$ for all $i$, leading to $A_i = 0$ for all $i$, yielding $\nabla_{\theta}J(\theta; q)=0$.

\textbf{HA-GRPO.} To address this limitation, a natural approach is to reduce the difficulty of the question $q$, which HA-GRPO methods achieve by adding a “hint” to the context. Let $z_q$ denote the reference solution to $q$, and $\mathcal{F}$ denote a method-specific hint-extraction mechanism (e.g., in QUESTA, $\mathcal{F}$ returns the first 50\% of tokens from $z_q$). The hinted question $q^+$, which replaces $q$, is devised as:
\begin{equation}
q^+ = \texttt{concat}[q, h_q], \quad\quad h_q := \mathcal{F}(q, z_q). \label{eqn-2}
\end{equation}

However, at inference time, the reference solution $z_{\overline{q}}$ for a given question $\overline{q}$ is naturally unavailable, so no hint $h_{\overline{q}}$ can be extracted and appended to $\overline{q}$ to form the hinted question $\overline{q}^+$. As a result, the model can no longer access the capability boost that hinting unlocked.

\textbf{Activation Steering.} Suppose the model $\pi_\theta$ consists of $L$ layers, and let $\mathbf{h}^{(l)}_t \in \mathbb{R}^{d'}$ denote the model's activation at layer $l \in \{1, ..., L\}$ (output of the decoder block) and token $t$. With a steering vector $\mathbf{v} \in \mathbb{R}^{d'}$ associated with a chosen layer $l$, $\mathbf{v}$ is added to the model's activation at that layer for every token $t$:
\begin{equation}
\tilde{\mathbf{h}}^{(l)}_t = \mathbf{h}^{(l)}_t + \mathbf{v}, \quad \forall t. 
\label{eqn-3}
\end{equation}
Then, the intervened $\tilde{\mathbf{h}}^{(l)}_t$ is propagated through the remaining layers $l+1, ..., L$ in place of $\mathbf{h}^{(l)}_t$. We denote the model with activation steering as $\pi_\theta^{\mathbf{v}}$ to distinguish it from $\pi_\theta$.

A central concern in activation steering is obtaining $\mathbf{v}$ that reliably captures the behavioral shift. Normally, $\mathbf{v}$ is computed offline and kept static – suitable when the target behavior is stable over time. In complex reasoning tasks, the target (hint-conditioned) behavior is entangled with a reference solution that varies across questions and must track the model as it evolves as training progresses.

\section{Activation Steering for Hinting}
\begin{figure}[!t]
\centering\includegraphics[width=\linewidth]{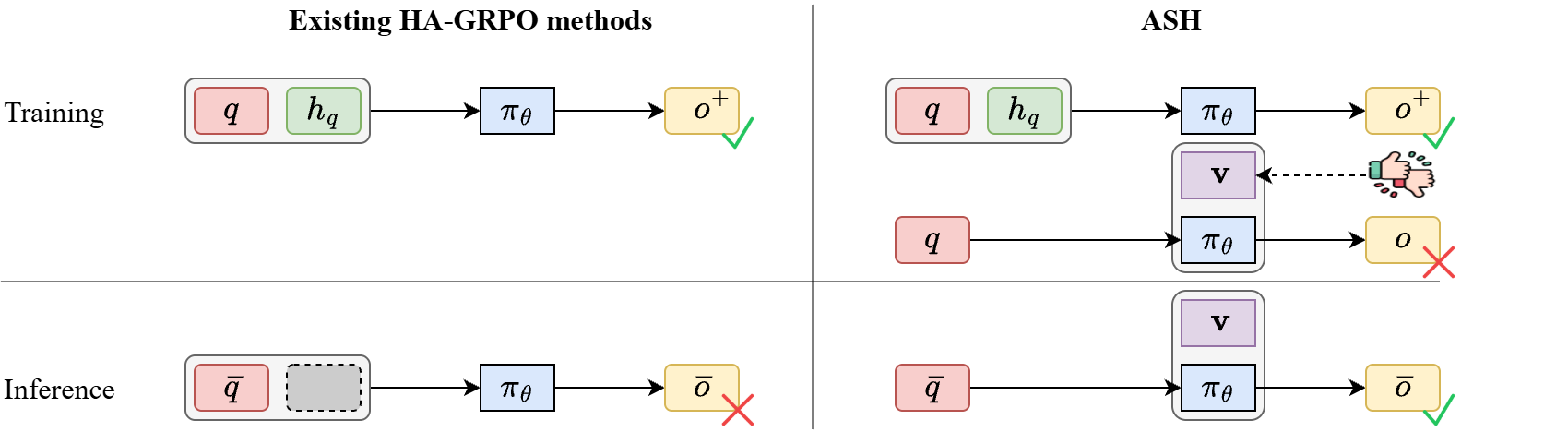}
\caption{A conceptual illustration of ASH, in comparison with existing HA-GRPO methods.}
\label{fig-2}
\end{figure}

We present ASH, a novel HA-GRPO method that operationalizes the view of hinting as steering model behavior. The design goal is to internalize the hint's behavioral shift in a trainable steering vector that remains available and can be used at inference time. Figure \ref{fig-2} gives a conceptual illustration of ASH.

\subsubsection*{Training}

First, as in other HA-GRPO methods, the hinted question $q^+$ is devised from the plain question $q$. Here, we adopt the simple mechanism $\mathcal{F}$ of QUESTA \citep{li2026questa}, which returns the first 50\% of tokens from $z_q$ (see examples in Appendix \ref{app-ex}). Subsequently, the model is presented with both the hinted question $q^+$ and the plain question $q$, and generates $\hat{G}$ responses to each version: \emph{hinted} $\{o_i^+ \sim \pi_\theta (\cdot|q^+)\}_{i=1, ..., \hat{G}}$ for $q^+$ and \emph{plain} $\{o_i \sim \pi_\theta (\cdot|q)\}_{i=1, ..., \hat{G}}$ for $q$. Then, we add the trainable steering vector $\mathbf{v}$ to the model's activation at a chosen layer $l$ (Equation \ref{eqn-3}) and optimize it to increase the differences in generation probabilities between \emph{hinted} responses relative to \emph{plain} responses, conditioned on $q$. Drawing on the analogy between our goal and that of Direct Preference Optimization (DPO) \citep{rafailov2023direct, xiao2024comprehensive}, we formulate a loss adapted from it:
\begin{equation}
\mathcal{L}(\mathbf{v}; q) = - \left[ \frac{1}{\hat{G}} \sum_{i=1}^{\hat{G}} \text{log}\text{ }\sigma \left( \beta\sum_{t=1}^{|o_i^+|}\text{log} \frac{\pi_\theta^{\mathbf{v}}(o_{i, t}^+|q, o_{i, <t}^+)}{\pi_\theta(o_{i, t}^+|q, o_{i, <t}^+)}-\beta\sum_{t=1}^{|o_i|}\text{log} \frac{\pi_\theta^{\mathbf{v}}(o_{i, t}|q, o_{i, <t})}{\pi_\theta(o_{i, t}|q, o_{i, <t})} \right) \right], 
\label{eqn-4}
\end{equation}
where $\beta$ is a scaling factor controlling how strongly the model with activation steering $\pi_\theta^{\mathbf{v}}$ is penalized for drifting too far from $\pi_\theta$. By minimizing this loss, the direction of $\mathbf{v}$ is accurately aligned with the question-agnostic component of the behavioral shift the hint induces. Note that we do not filter $\hat{G}$ responses to the hinted question $q^+$ and $\hat{G}$ responses to the plain question $q$ based on their correctness at this stage, as we strive to isolate the behavioral shift in the model, independent of whether the final answers are correct.

Once the steering vector $\mathbf{v}$ has been trained, the model itself is then trained with GRPO as in other HA-GRPO methods, but on both the hinted question $q^+$ and the plain question $q$, and what it has generated, i.e., \emph{hinted} $\{o_i^+ \sim \pi_\theta (\cdot|q^+)\}_{i=1, ..., \hat{G}}$ for $q^+$ and \emph{plain} $\{o_i \sim \pi_\theta (\cdot|q)\}_{i=1, ..., \hat{G}}$ for $q$. Denote their rewards as $\{R_i^+\}_{i=1, ..., \hat{G}}$ and $\{R_i\}_{i=1, ..., \hat{G}}$, respectively, the advantages of each $\hat{G}$ responses are computed via:
\begin{equation*}
\{A_i^+ = \frac{R_i^+ - \text{mean}(\{R_i^+\}_{i=1, ..., \hat{G}})}{\text{std}(\{R_i^+\}_{i=1, ..., \hat{G}})+\varepsilon}\}_{i=1, ..., \hat{G}} \text{ and } \{A_i = \frac{R_i - \text{mean}(\{R_i\}_{i=1, ..., \hat{G}})}{\text{std}(\{R_i\}_{i=1, ..., \hat{G}})+\varepsilon}\}_{i=1, ..., \hat{G}}. \\
\end{equation*}
Following that, the model is updated by minimizing the sum of $J(\theta; q^+)$ and $J(\theta; q)$:
\begin{gather}
J(\theta; q^+) + J(\theta; q) = - \left[ \frac{1}{\hat{G}} \sum_{i=1}^{\hat{G}} \frac{1}{|o_i^+|} \sum_{t=1}^{|o_i^+|} \left( r_\theta(o_{i, t}^+)A_i^+ \right) \right] - \left[ \frac{1}{\hat{G}} \sum_{i=1}^{\hat{G}} \frac{1}{|o_i|} \sum_{t=1}^{|o_i|} \left( r_\theta(o_{i, t})A_i \right) \right], \nonumber \\
\text{where} \quad r_\theta(o_{i, t}^+) = \frac{\pi_\theta(o_{i, t}^+ | q^+, o_{i, < t}^+)}{[\pi_\theta(o_{i, t}^+ | q^+, o_{i, < t}^+)]_{\text{not\_updated}}} \text{ and } r_\theta(o_{i, t}) = \frac{\pi_\theta(o_{i, t} | q, o_{i, < t})}{[\pi_\theta(o_{i, t} | q, o_{i, < t})]_{\text{not\_updated}}}. 
\label{eqn-5}
\end{gather}
Note that the model is trained on both $q^+$ and $q$ jointly, rather than on $q^+$ alone. This is possible without interference, since $J(\theta; q^+)$ and $J(\theta; q)$ are computed from independently estimated advantages and simply summed, so they do not compete with each other. Doing so makes fuller use of the already-generated responses for both $q^+$ and $q$, and additionally acts as a form of \textbf{regularization} that reduces the model's reliance on the hint.

\textbf{Co-Training Ordering.} These two stages together form a single training step, and their order is crucial. The steering vector $\mathbf{v}$ must be optimized before the model is updated since optimizing $\mathbf{v}$ afterward would leave it fit to a stale model, misaligned with the one that actually generated the \emph{hinted} $\{o_i^+ \sim \pi_\theta (\cdot|q^+)\}_{i=1, ..., \hat{G}}$ and \emph{plain} $\{o_i \sim \pi_\theta (\cdot|q)\}_{i=1, ..., \hat{G}}$ in Equation \ref{eqn-4}. Such misalignment causes $\mathbf{v}$ to steer model behavior in a direction that no longer corresponds to the hint's true behavioral shift. This order incurs no interference, as optimizing $\mathbf{v}$ leaves the model untouched. Moreover, because $\mathbf{v}$ is re-optimized at every training step, it tracks the model's growth as training progresses, keeping activation steering faithful to the current model rather than to an outdated one. To preserve this correspondence at termination, the model is not updated at the final training step. Algorithm \ref{alg-1} depicts the training procedure. Empirical training dynamics are shown in Appendix \ref{app-logs}.

\begin{algorithm}[!t]
\caption{ASH’s training procedure.}\label{alg-1}
\begin{algorithmic}[1]
\Statex \textbf{Input:} A model $\pi_\theta$; training data $\mathcal{D}$; number of training steps $T$ and the step size $\eta$
\Statex \textbf{Output:} The trained steering vector $\mathbf{v}^*$; the trained parameters $\theta^*$
\State $\mathbf{v} \gets \mathbf{0}$ \Comment{initialize the steering vector}
\For{training step $= 1$ to $T$}
\State sample the question $q$ and the reference solution $z_q$ from $\mathcal{D}$
\State $q^+ = \texttt{concat}[q, h_q], \quad\quad h_q := \mathcal{F}(q, z_q)$ \Comment{Equation \ref{eqn-2}}
\State obtain \emph{hinted} $\{o_i^+ \sim \pi_\theta (\cdot|q^+)\}_{i=1, ..., \hat{G}}$ and \emph{plain} $\{o_i \sim \pi_\theta (\cdot|q)\}_{i=1, ..., \hat{G}}$
\State compute $\mathcal{L}(\mathbf{v}; q)$ using $\{o_i^+\}_{i=1, ..., \hat{G}}$ and $\{o_i\}_{i=1, ..., \hat{G}}$ \Comment{Equation \ref{eqn-4}}
\State $\mathbf{v} \gets \mathbf{v} - \eta\nabla_{\mathbf{v}} \mathcal{L}(\mathbf{v}; q)$
\If{training step $< T$}
    \State compute $\{A_i^+\}_{i=1, ..., \hat{G}}$ and $\{A_i\}_{i=1, ..., \hat{G}}$
    \State compute $J(\theta; q^+) + J(\theta; q)$ \Comment{Equation \ref{eqn-5}}
    \State $\theta \gets \theta - \eta\nabla_\theta \left[ J(\theta; q^+) + J(\theta; q) \right]$
\EndIf
\EndFor
\State \Return $\mathbf{v}^*$, $\theta^*$
\end{algorithmic}
\end{algorithm}

\subsubsection*{Inference}
After all, given a question $\overline{q}$, the trained steering vector $\mathbf{v}^*$ is directly injected into the model's activation at the same layer $l$ (Equation \ref{eqn-3}) without any accompanying hint $h_{\overline{q}}$. This facilitates the model with activation steering $\pi_{\theta^*}^{\mathbf{v}^*}$ to reproduce the hint-conditioned behavior on $\overline{q}$, thus precisely satisfying the design goal of ASH.

\section{Theoretical Analysis}
We provide a theoretical analysis establishing how ASH bounds and shrinks the performance gap caused by the training-inference context mismatch. The complete proof is deferred to Appendix \ref{app-proof}.
\begin{theorem}[Performance Gap Bound with Activation Steering]
\label{theorem-1}
Given a training question $q$ and the hinted question $q^+$ devised from it, and a bounded reward function $\mathcal{R}: \mathcal{O} \to \{0, 1\}$ over generated responses $o \in \mathcal{O}$, the expected performance gap is upper-bounded by:
\begin{equation}
| \mathbb{E}_{o \sim \pi_\theta(\cdot | q^+)}[\mathcal{R}(o)] - \mathbb{E}_{o \sim \pi_\theta^{\mathbf{v}}(\cdot | q)}[\mathcal{R}(o)] | \le \sqrt{2 D_{\text{KL}}(\pi_\theta(\cdot | q^+) \parallel \pi_\theta^{\mathbf{v}}(\cdot | q))}. 
\label{eqn-6}
\end{equation}
\end{theorem}
Intuitively, minimizing $\mathcal{L}(\mathbf{v}; q)$ increases the likelihood of \emph{hinted} responses relative to \emph{plain} responses, under $q$, which we expect to move $\pi_\theta^{\mathbf{v}}(\cdot | q)$ toward $\pi_\theta(\cdot | q^+)$ and thus reduce the divergence on the right-hand side and tighten the upper bound on the performance gap.

\section{Experiments}
\label{sec-exps}
In this section, we systematically conduct a series of experiments to empirically evaluate our proposed ASH in expanding the reasoning capability of LLMs. Our evaluation is structured around the following key research questions (RQs):
\begin{itemize}[topsep=0em, leftmargin=*, noitemsep]
\item \textbf{RQ1 - Mechanistic Validity:} Does the regularization of joint training on both hinted questions and plain questions help, and to what extent? Importantly, to what extent does activation steering recover the capability boost that hinting unlocked?
\item \textbf{RQ2 - Effectiveness:} Does the design of ASH translate into tangible improvements in challenging reasoning benchmarks over standard GRPO and other HA-GRPO methods?
\item \textbf{RQ3 - Isolated Contribution:} How much does activation steering contribute to?
\item \textbf{RQ4 - Cross-Domain Transferability:} Does the trained steering vector, which was fit on mathematics, transfer effectively to other reasoning domains?
\end{itemize}

\subsection{Setup}

\textbf{Training Data.} We employ the challenging OpenR1-MATH-220k dataset \citep{openr1}, comprising 220k mathematics questions collected from diverse sources, each paired with a reference solution generated by DeepSeek-R1 \citep{guo2025deepseek}. Following prior work \citep{liao2026self, xia2026}, we curate a subset of 15k questions and utilize it as training data (Appendix \ref{app-hyperparameters}).

\begin{figure}[!t]\centering
\begin{subfigure}{0.49\textwidth}\centering
\includegraphics[width=0.866\linewidth]{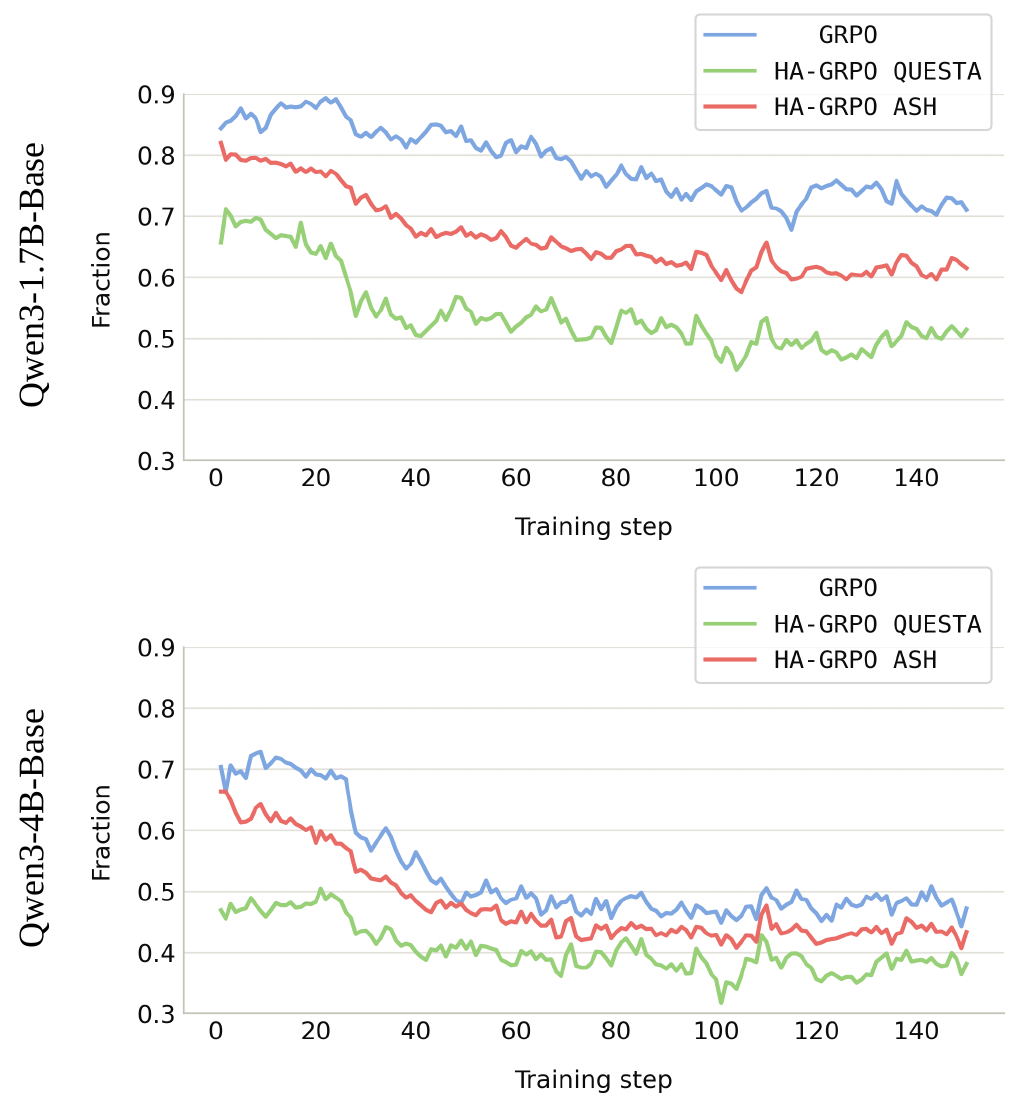}
\caption{The fraction of too-difficult questions per batch over the course of training.}
\label{fig-3a}
\end{subfigure}
\hfill
\begin{subfigure}{0.49\textwidth}
\centering
\includegraphics[width=0.866\linewidth]{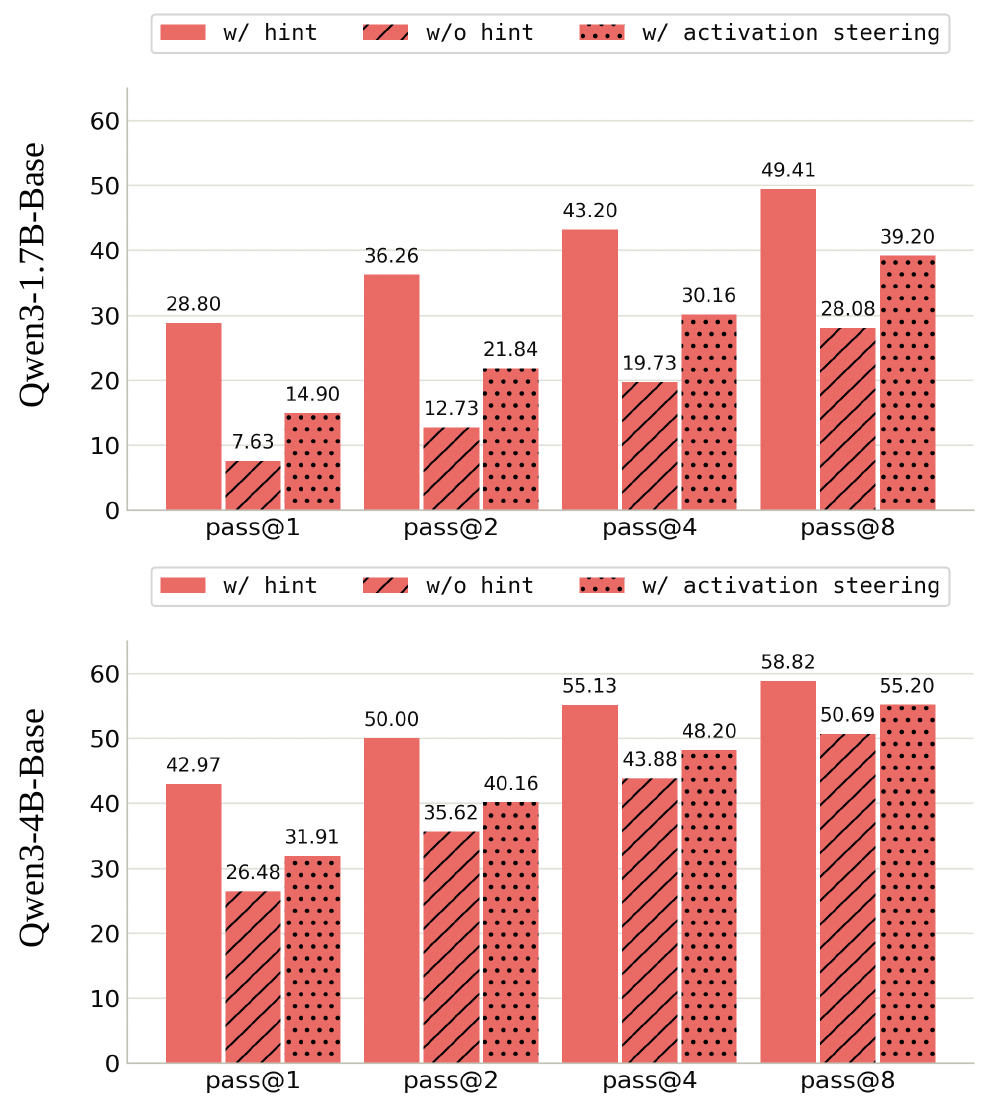}
\caption{pass@$k$ of ASH-trained models, with and without the hint in the questions, and with activation steering.}
\label{fig-3b}
\end{subfigure}
\caption{The impact of the training-inference context mismatch. (a) ASH substantially lessens the fraction of too-difficult questions per batch over the course of training, relative to GRPO. (b) pass@$k$ of ASH-trained models drops sharply without the hint at inference time. However, activation steering significantly restores it.}
\end{figure}

\textbf{Base LLMs.} We evaluate four LLMs of varying scales: Qwen3-1.7B-Base and Qwen3-4B-Base \citep{yang2025qwen3}, and Llama-3.2-1B-Instruct and Llama-3.2-3B-Instruct \citep{grattafiori2024llama}.

\textbf{Baselines.} We compare ASH against standard GRPO and representative HA-GRPO methods, including QUESTA, SAGE, and HILL.

\textbf{Implementation Details.} For baselines, each model generates 8 responses (i.e., $G=8$) per training question, with an instruction prepended to form the input prompt: “\texttt{Please reason step by step, and put your final answer within} $\verb|\|\texttt{boxed}\{\}$”. For ASH, each model instead generates 4 responses to the hinted question and 4 responses to the plain question (i.e., $\hat{G}=4$), matching the generation budget of baselines, preserving an \underline{apples-to-apples} comparison. Other key training hyperparameters are provided in Appendix \ref{app-hyperparameters}.

\textbf{Testing Benchmarks.} We evaluate on six benchmarks across two categories: competition-level mathematics with AMC \citep{li2024numinamath}, OlympiadBench \citep{he2024olympiadbench}, AIME24, AIME25, and out-of-distribution with Minerva \citep{lewkowycz2022}, GPQA-Diamond \citep{rein2024gpqa}.

\textbf{Metric.} We report pass@$k$ \citep{chen2021} with $k$ up to 8, which meaningfully reflects the expansion in models' reasoning capability. We provide a discussion of this metric in Appendix \ref{app-metric}.

\subsection{RQ1 - Mechanistic Validity}

With the described setup, we first report the impact of the training-inference context mismatch. As shown in Figure \ref{fig-3a}, ASH substantially lessens the fraction of too-difficult questions per batch over the course of training, relative to GRPO, aligned with expectations. We then evaluate ASH-trained models on training data, with and without the hint in the questions. As shown in Figure \ref{fig-3b}, pass@$k$ drops sharply without the hint at inference time; for example, pass@8 falls by \textbf{21.33}\% and \textbf{08.13}\% for Qwen3-1.7B-Base and Qwen3-4B-Base, respectively.

\textbf{The regularization helps.} ASH retains core properties of HA-GRPO methods like QUESTA, while the regularization of joint training on both hinted questions and plain questions positions it in the \textit{middle} of GRPO and QUESTA. This regularization reduces the extent to which models trained with ASH rely on the hint as a crutch during training and generalizes better to the hint-free setting during inference, reflected in narrowed gaps: \textbf{21.33}\% and \textbf{08.13}\%, compared to \textbf{27.34}\% and \textbf{10.83}\%.

\textbf{Moreover, activation steering recovers about half of the capability boost that hinting unlocked.} Figure \ref{fig-3b} also shows that activation steering significantly restores pass@$k$ of ASH-trained models, raising pass@8 back by \textbf{11.11}\% and \textbf{04.51}\% for Qwen3-1.7B-Base and Qwen3-4B-Base, i.e., about half of the gaps. This supports the idea that the steering vector captured the question-agnostic component of the behavioral shift the hint induces.

\begin{table}[!t]
\centering\scriptsize
\caption{
pass@8 of models on six testing benchmarks. Best results are in \textbf{bold}. 
}
\label{tab-1}
\setlength\tabcolsep{0pt}\begin{tabular*}{\linewidth}{@{\extracolsep{\fill}} lrlccccc@{\hspace{5pt}}|ccc }
\midrule
Model &Method & &AMC &OlympiadBench &AIME24 &AIME25 &Average &Minerva &GPQA-Diamond &Average \\
\midrule
\multirow[t]{4}{*}{Qwen3-1.7B-Base}       &Base   &-    &55.77 &46.48 &10.24 &13.84 &31.58 &39.10 &39.79 &39.45 \\
                                          &GRPO   &-    &56.06 &50.14 &21.44 &14.99 &35.66 &48.99 &61.10 &55.05 \\
& \multirow[t]{4}{*}{HA-GRPO}             &QUESTA       &56.30 &51.04 &17.30 &14.72 &34.84 &47.69 &69.78 &58.73 \\
&                                         &SAGE         &58.65 &51.56 &20.24 &14.88 &36.33 &48.77 &63.54 &56.16 \\
&                                         &HILL         &59.13 &52.75 &21.72 &16.51 &37.53 &49.23 &64.76 &57.00 \\
&                                         &\cellcolor{gred!20}ASH &\cellcolor{gred!20}\textbf{60.73} &\cellcolor{gred!20}\textbf{54.59} &\cellcolor{gred!20}\textbf{23.79} &\cellcolor{gred!20}\textbf{18.91} &\cellcolor{gred!20}\textbf{39.50} &\cellcolor{gred!20}\textbf{51.52} &\cellcolor{gred!20}\textbf{70.15} &\cellcolor{gred!20}\textbf{60.84} \\
\midrule
\multirow[t]{4}{*}{Qwen3-4B-Base}         &Base   &-    &69.11 &59.35 &29.27 &25.75 &45.87 &53.00 &52.36 &52.68 \\
                                          &GRPO   &-    &71.67 &63.83 &34.24 &32.43 &50.54 &60.37 &73.36 &66.86 \\
& \multirow[t]{4}{*}{HA-GRPO}             &QUESTA       &73.75 &64.49 &32.00 &29.32 &49.89 &58.61 &75.72 &67.16 \\
&                                         &SAGE         &76.25 &64.75 &33.61 &32.15 &51.69 &60.28 &75.75 &68.02 \\
&                                         &HILL         &77.10 &65.85 &36.40 &34.95 &53.58 &60.49 &76.19 &68.34 \\
&                                         &\cellcolor{gred!20}ASH &\cellcolor{gred!20}\textbf{79.21} &\cellcolor{gred!20}\textbf{67.71} &\cellcolor{gred!20}\textbf{39.11} &\cellcolor{gred!20}\textbf{37.31} &\cellcolor{gred!20}\textbf{55.83} &\cellcolor{gred!20}\textbf{62.65} &\cellcolor{gred!20}\textbf{78.28} &\cellcolor{gred!20}\textbf{70.47} \\
\midrule
\multirow[t]{4}{*}{Llama-3.2-1B-Instruct} &Base   &-    &30.77 &16.80 &07.25 &00.00 &13.71 &12.27 &10.80 &11.54 \\
                                          &GRPO   &-    &34.75 &21.15 &11.97 &00.00 &16.97 &20.07 &13.85 &16.96 \\
& \multirow[t]{4}{*}{HA-GRPO}             &QUESTA       &38.06 &24.00 &10.86 &00.00 &18.23 &18.20 &16.10 &17.15 \\
&                                         &SAGE         &40.66 &24.56 &11.56 &02.24 &19.75 &19.21 &14.85 &17.03 \\
&                                         &HILL         &41.02 &25.14 &13.10 &03.83 &20.77 &20.91 &15.52 &18.21 \\
&                                         &\cellcolor{gred!20}ASH &\cellcolor{gred!20}\textbf{43.57} &\cellcolor{gred!20}\textbf{27.18} &\cellcolor{gred!20}\textbf{14.33} &\cellcolor{gred!20}\textbf{05.45} &\cellcolor{gred!20}\textbf{22.63} &\cellcolor{gred!20}\textbf{21.05} &\cellcolor{gred!20}\textbf{20.97} &\cellcolor{gred!20}\textbf{21.01} \\
\midrule
\multirow[t]{4}{*}{Llama-3.2-3B-Instruct} &Base   &-    &49.23 &38.09 &19.72 &00.83 &26.97 &36.87 &26.34 &31.60 \\
                                          &GRPO   &-    &51.58 &38.95 &22.46 &04.17 &29.29 &39.14 &26.34 &32.74 \\
& \multirow[t]{4}{*}{HA-GRPO}             &QUESTA       &54.40 &40.38 &19.35 &04.81 &29.73 &38.12 &28.11 &33.11 \\
&                                         &SAGE         &55.60 &43.87 &21.16 &05.58 &31.55 &38.37 &28.29 &33.33 \\
&                                         &HILL         &57.06 &44.80 &24.24 &06.06 &33.04 &39.88 &28.31 &34.09 \\
&                                         &\cellcolor{gred!20}ASH &\cellcolor{gred!20}\textbf{61.19} &\cellcolor{gred!20}\textbf{46.25} &\cellcolor{gred!20}\textbf{25.99} &\cellcolor{gred!20}\textbf{07.34} &\cellcolor{gred!20}\textbf{35.19} &\cellcolor{gred!20}\textbf{40.91} &\cellcolor{gred!20}\textbf{31.50} &\cellcolor{gred!20}\textbf{36.20} \\
\midrule
\end{tabular*}
\end{table}

\subsection{RQ2 - Effectiveness}

Having achieved mechanistic validity, we next test whether ASH's design produces concrete improvements in pass@$k$, evaluated on both competition-level mathematics benchmarks and out-of-distribution benchmarks. Table \ref{tab-1} reports pass@8 of models on six testing benchmarks. 

\textbf{ASH improves pass@8 across four models and all six benchmarks.} Compared to GRPO, ASH achieves comprehensive and significant performance gains across all base models. On average, it improves pass@8 by \textbf{3.84}\%/\textbf{5.29}\% on competition-level mathematics and \textbf{5.79}\%/\textbf{3.61}\% on out-of-distribution benchmarks for Qwen3-1.7B/4B-Base, and by \textbf{5.66}\%/\textbf{5.90}\% on competition-level mathematics and \textbf{4.05}\%/\textbf{3.46}\% on out-of-distribution benchmarks for Llama-3.2-1B/3B-Instruct.

\textbf{ASH surpasses existing HA-GRPO methods.} Relative to GRPO, QUESTA yields inconsistent performance gains across benchmarks, falling below GRPO on the challenging AIME24/AIME25 and Minerva. HILL, which employs a more advanced hint-extraction mechanism, achieves consistent gains, but at nearly double the computational cost (Appendix \ref{app-hyperparameters}). ASH, in contrast, achieves this consistency with only manageable computational overhead. Over these methods, ASH delivers average gains of \textbf{2.60}\%/\textbf{2.21}\% for Qwen3-1.7B/4B-Base, and \textbf{2.17}\%/\textbf{2.14}\% for Llama-3.2-1B/3B-Instruct. These gains split into \textbf{1.97}\%/\textbf{2.25}\% on competition-level mathematics and \textbf{3.84}\%/\textbf{2.13}\% on out-of-distribution benchmarks for Qwen3-1.7B/4B-Base, and \textbf{1.86}\%/\textbf{2.15}\% on competition-level mathematics and \textbf{2.80}\%/\textbf{2.11}\% on out-of-distribution benchmarks for Llama-3.2-1B/3B-Instruct.

These results demonstrate that the training-inference context mismatch limits performance gains achievable by existing HA-GRPO methods. By mitigating it, ASH substantially expands the reasoning capability of LLMs, with gains that generalize consistently across model families and scales.

\subsection{RQ3 - Isolated Contribution}

To quantify how much of ASH's performance gains come from activation steering, we evaluate ASH-trained models with and without applying it. The remaining improvements over GRPO can be attributed to regularization from joint training. Table \ref{tab-2} reports pass@8 on six testing benchmarks.

\textbf{Applying activation steering plays a pivotal role in ASH's performance gains.} In pass@8, applying it accounts for \textbf{3.78}\%/\textbf{4.55}\% on competition-level mathematics and \textbf{1.72}\%/\textbf{2.66}\% on out-of-distribution benchmarks for Qwen3-1.7B/4B-Base, and for \textbf{3.70}\%/\textbf{4.66}\% on competition-level mathematics and \textbf{3.13}\%/\textbf{2.92}\% on out-of-distribution benchmarks for Llama-3.2-1B/3B-Instruct. Compared to GRPO, on average, applying activation steering contributes to approximately 68.9\%/83.1\% for Qwen3-1.7B/4B-Base, and 68.6\%/80.2\% for Llama-3.2-1B/3B-Instruct.

These results imply the importance of the trained steering vector, which indeed genuinely steers model behavior toward a higher-quality reasoning space.

\begin{table}[!t]
\centering\scriptsize
\caption{
pass@8 of ASH-trained models on six testing benchmarks. 
}
\label{tab-2}
\setlength\tabcolsep{0pt}\begin{tabular*}{\linewidth}{@{\extracolsep{\fill}} llccccc@{\hspace{5pt}}|ccc }
\midrule
Model & &AMC &OlympiadBench &AIME24 &AIME25 &Average &Minerva &GPQA-Diamond &Average \\
\midrule
\multirow[t]{4}{*}{Qwen3-1.7B-Base}       &w/o activation steering                              &57.35 &51.51 &19.28 &14.74 &35.72 &48.31 &69.92 &59.12 \\
                                          &\cellcolor{gred!20}w/\phantom{o} activation steering &\cellcolor{gred!20}60.73 &\cellcolor{gred!20}54.59 &\cellcolor{gred!20}23.79 &\cellcolor{gred!20}18.91 &\cellcolor{gred!20}39.50 &\cellcolor{gred!20}51.52 &\cellcolor{gred!20}70.15 &\cellcolor{gred!20}60.84 \\
\midrule
\multirow[t]{4}{*}{Qwen3-4B-Base}         &w/o activation steering                              &75.15 &64.68 &33.45 &31.82 &51.28 &59.88 &75.74 &67.81 \\
                                          &\cellcolor{gred!20}w/\phantom{o} activation steering &\cellcolor{gred!20}79.21 &\cellcolor{gred!20}67.71 &\cellcolor{gred!20}39.11 &\cellcolor{gred!20}37.31 &\cellcolor{gred!20}55.83 &\cellcolor{gred!20}62.65 &\cellcolor{gred!20}78.28 &\cellcolor{gred!20}70.47 \\
\midrule
\multirow[t]{4}{*}{Llama-3.2-1B-Instruct} &w/o activation steering                              &39.31 &24.38 &11.22 &00.80 &18.93 &18.81 &16.94 &17.88 \\
                                          &\cellcolor{gred!20}w/\phantom{o} activation steering &\cellcolor{gred!20}43.57 &\cellcolor{gred!20}27.18 &\cellcolor{gred!20}14.33 &\cellcolor{gred!20}05.45 &\cellcolor{gred!20}22.63 &\cellcolor{gred!20}21.05 &\cellcolor{gred!20}20.97 &\cellcolor{gred!20}21.01 \\
\midrule
\multirow[t]{4}{*}{Llama-3.2-3B-Instruct} &w/o activation steering                              &55.21 &41.18 &20.47 &05.27 &30.53 &38.32 &28.25 &33.28 \\
                                          &\cellcolor{gred!20}w/\phantom{o} activation steering &\cellcolor{gred!20}61.19 &\cellcolor{gred!20}46.25 &\cellcolor{gred!20}25.99 &\cellcolor{gred!20}07.34 &\cellcolor{gred!20}35.19 &\cellcolor{gred!20}40.91 &\cellcolor{gred!20}31.50 &\cellcolor{gred!20}36.20 \\
\midrule
\end{tabular*}
\end{table}
\subsection{RQ4 - Cross-Domain Transferability}
\begin{wraptable}[13]{r}{0.53\textwidth}
\centering\scriptsize
\vspace{-13.1pt}
\caption{
pass@8 of ASH-trained models on the LogiQA and LiveCodeBench-v5 benchmarks. 
}
\label{tab-3}
\setlength\tabcolsep{0pt}\begin{tabular*}{\linewidth}{@{\extracolsep{\fill}} llcc }
\midrule
Model                                       &                                                      & LogiQA                    & LiveCodeBench-v5         \\
\midrule
\multirow[t]{2}{*}{Qwen3-1.7B-Base}         & w/o activation steering                              & 79.57                     & 16.13                    \\
                                            & \cellcolor{gred!20}w/\phantom{o} activation steering & \cellcolor{gred!20}88.69  & \cellcolor{gred!20}18.77 \\
\midrule
\multirow[t]{2}{*}{Qwen3-4B-Base}           & w/o activation steering                              & 81.87                     & 30.47                    \\
                                            & \cellcolor{gred!20}w/\phantom{o} activation steering & \cellcolor{gred!20}89.10  & \cellcolor{gred!20}33.47 \\
\midrule
\multirow[t]{2}{*}{Llama-3.2-1B-Instruct}   & w/o activation steering                              & 54.84                     & 05.74                    \\
                                            & \cellcolor{gred!20}w/\phantom{o} activation steering & \cellcolor{gred!20}59.10  & \cellcolor{gred!20}08.20 \\
\midrule
\multirow[t]{2}{*}{Llama-3.2-3B-Instruct}   & w/o activation steering                              & 68.20                     & 19.35                    \\
                                            & \cellcolor{gred!20}w/\phantom{o} activation steering & \cellcolor{gred!20}71.94  & \cellcolor{gred!20}20.89 \\
\midrule
\end{tabular*}
\end{wraptable}
We investigate whether the trained steering vector fit on mathematics can transfer to other reasoning domains. Thus, we examine ASH-trained models with and without applying activation steering on two domains beyond mathematics: logical question answering and coding, using LogiQA benchmark and LiveCodeBench-v5 benchmark \citep{liu2020logiqa, jain2025}, respectively. Table \ref{tab-3} reports pass@8 on these two benchmarks.

\textbf{The trained steering vector can transfer effectively to other reasoning domains.} On top of ASH-trained models, activation steering brings salient gains in both logical question answering and coding, achieving transferability without any further training. On LogiQA and LiveCodeBench-v5 benchmarks, it improves pass@8 for Qwen3-1.7B/4B-Base by \textbf{9.12}\%/\textbf{7.23}\% and \textbf{2.64}\%/\textbf{3.00}\%, and for Llama-3.2-1B/3B-Instruct by \textbf{4.26}\%/\textbf{3.74}\% and \textbf{2.46}\%/\textbf{1.54}\%.

These results, again, indicate that the trained steering vector from ASH genuinely steers model behavior toward a higher-quality reasoning space and is transferable across reasoning domains.

\section{Limitations}
One potential limitation of ASH is the computational overhead of training the steering vector. This overhead is modest in practice for two reasons. First, the implicit rewards in Equation \ref{eqn-4} require no additional generation: they are computed by parallel prefilling over the \emph{hinted} and \emph{plain} responses that have already been generated. Second, we compute $\mathcal{L}(\mathbf{v}; q)$ only over index-wise pairs of responses rather than all possible combinations, which is sufficient for training the steering vector (Appendix \ref{app-logs}). As a result, ASH requires about 1.5× the training time of GRPO, substantially less than the 2× required by SAGE and HILL (Appendix \ref{app-hyperparameters}), while achieving higher effectiveness.

\section{Conclusion}
We introduced ASH, a novel HA-GRPO method that mitigates the training-inference context mismatch by reinterpreting hinting as a means of steering model behavior. Across four LLMs, ASH consistently improves pass@$k$ on all six benchmarks and establishes a steering vector that can transfer effectively across reasoning domains. Moreover, ASH is orthogonal to, and compatible with, other HA-GRPO methods such as SAGE and HILL. Finally, ASH employs a single global steering vector; enriching its capacity with task- or question-specific steering vectors is promising for future work.

\section*{AI Use Statement}
The authors used a generative AI tool to improve the grammar and readability of portions of the manuscript and to assist with coding. All AI-generated suggestions and code were reviewed, verified, and edited by the authors. The authors take full responsibility for the final content of the manuscript, including the code and analyses.

% \clearpage
\bibliography{iclr2027_conference}
\bibliographystyle{iclr2027_conference}

\clearpage
\appendix
\section{Creating Hinted Questions}
\label{app-ex}
We provide an example of how hinted questions are devised from plain questions. Note that we adopt the simple hint-extraction mechanism $\mathcal{F}$ of QUESTA \citep{li2026questa}.
\\\\
\textbf{Example:}
\begin{tcolorbox}[left=5pt, right=5pt, top=5pt, bottom=5pt, fonttitle=\normalsize, colback=red!5, colframe=red!50, title={Plain Question}]
\begin{minipage}{\textwidth}
\normalsize
Concatenate the natural number $N$ to the right of every natural number (for example: appending 2 to 35 results in 352). If the new number can always be divided by $N$, then $N$ is called a magic number. Among the natural numbers less than 130, the number of magic numbers is $\underline{\hspace{2cm}}$
\end{minipage}
\end{tcolorbox}
\begin{tcolorbox}[left=5pt, right=5pt, top=5pt, bottom=5pt, fonttitle=\normalsize, colback=Green!5, colframe=Green!50, title={Reference Solution}]
\begin{minipage}{\textwidth}
\normalsize
To determine the number of magic numbers less than 130, we need to identify natural numbers $N$ such that appending $N$ to any natural number results in a number divisible by $N$. This requires $N$ to divide $10^d$, where $d$ is the number of digits in $N$.

\begin{enumerate}
    \item \textbf{For 1-digit numbers ($d=1$):}
    \begin{itemize}
        \item Divisors of $10^1 = 10$ that are 1-digit numbers: $1, 2, 5$.
    \end{itemize}

    \item \textbf{For 2-digit numbers ($d=2$):}
    \begin{itemize}
        \item Divisors of $10^2 = 100$ that are 2-digit numbers: $10, 20, 25, 50$.
    \end{itemize}

    \item \textbf{For 3-digit numbers ($d=3$):}
    \begin{itemize}
        \item Divisors of $10^3 = 1000$ that are 3-digit numbers less than 130: $100, 125$.
    \end{itemize}
\end{enumerate}

Listing all magic numbers less than 130:
\begin{itemize}
    \item 1-digit: $1, 2, 5$ (3 numbers)
    \item 2-digit: $10, 20, 25, 50$ (4 numbers)
    \item 3-digit: $100, 125$ (2 numbers)
\end{itemize}

Summing these, we get a total of $3 + 4 + 2 = 9$ magic numbers.

Thus, the number of magic numbers less than 130 is $\boxed{9}$.
\end{minipage}
\end{tcolorbox}
\begin{tcolorbox}[left=5pt, right=5pt, top=5pt, bottom=5pt, fonttitle=\normalsize, colback=red!5, colframe=red!50, title={Hinted Question}]
\begin{minipage}{\textwidth}
\normalsize
Concatenate the natural number $N$ to the right of every natural number (for example: appending 2 to 35 results in 352). If the new number can always be divided by $N$, then $N$ is called a magic number. Among the natural numbers less than 130, the number of magic numbers is $\underline{\hspace{2cm}}$
\\\\
\textbf{The following text is the beginning part of the solution, which you can refer to for solving the problem:}

To determine the number of magic numbers less than 130, we need to identify natural numbers $N$ such that appending $N$ to any natural number results in a number divisible by $N$. This requires $N$ to divide $10^d$, where $d$ is the number of digits in $N$.

\begin{enumerate}
    \item \textbf{For 1-digit numbers ($d=1$):}
    \begin{itemize}
        \item Divisors of $10^1 = 10$ that are 1-digit numbers: $1, 2, 5$.
    \end{itemize}

    \item \textbf{For 2-digit numbers ($d=2$):}
    \begin{itemize}
        \item Divisors of $10^2 = 100$ that are 2-digit numbers: $10, 20, 25, 50$.
    \end{itemize}

    \item \textbf{For 3-digit numbers ($d=3$):}
    \begin{itemize}
        \item Divisors of $10^3 = 1000$ that are 3-digit numbers less than 130: $100, 125$.
    \end{itemize}
\end{enumerate}
\end{minipage}
\end{tcolorbox}

\section{Proof of Theorem \ref{theorem-1}}
\label{app-proof}
\begin{proof}
\phantom{}

The absolute difference in expected rewards (expected performance gap) is written as: 
$$| \mathbb{E}_{o \sim \pi_\theta(\cdot | q^+)}[\mathcal{R}(o)] - \mathbb{E}_{o \sim \pi_\theta^{\mathbf{v}}(\cdot | q)}[\mathcal{R}(o)] | = \left| \sum_{o \in \mathcal{O}} \mathcal{R}(o)(\pi_\theta(o | q^+) - \pi_\theta^{\mathbf{v}}(o | q)) \right|$$
Since the reward function $\mathcal{R}$ is bounded such that $0 \le \mathcal{R}(o) \le 1$ for all $o \in \mathcal{O}$: 
$$\left| \sum_{o \in \mathcal{O}} \mathcal{R}(o)(\pi_\theta(o | q^+) - \pi_\theta^{\mathbf{v}}(o | q)) \right| \le \sum_{o \in \mathcal{O}} \mathcal{R}(o) | \pi_\theta(o | q^+) - \pi_\theta^{\mathbf{v}}(o | q) | \le \sum_{o \in \mathcal{O}} | \pi_\theta(o | q^+) - \pi_\theta^{\mathbf{v}}(o | q) |$$
By definition, the Total Variation (TV) distance between two distributions is: 
$$D_{\text{TV}}(\pi_\theta(\cdot | q^+), \pi_\theta^{\mathbf{v}}(\cdot | q)) = \frac{1}{2} \sum_{o \in \mathcal{O}} | \pi_\theta(o | q^+) - \pi_\theta^{\mathbf{v}}(o | q) |$$
Substituting this into the inequality gives: 
$$| \mathbb{E}_{o \sim \pi_\theta(\cdot | q^+)}[\mathcal{R}(o)] - \mathbb{E}_{o \sim \pi_\theta^{\mathbf{v}}(\cdot | q)}[\mathcal{R}(o)] | \le 2 D_{\text{TV}}(\pi_\theta(\cdot | q^+), \pi_\theta^{\mathbf{v}}(\cdot | q))$$
Applying Pinsker's Inequality, which relates the Total Variation distance to the KL divergence: 
$$D_{\text{TV}}(\pi_\theta(\cdot | q^+), \pi_\theta^{\mathbf{v}}(\cdot | q)) \le \sqrt{\frac{1}{2} D_{\text{KL}}(\pi_\theta(\cdot | q^+) \parallel \pi_\theta^{\mathbf{v}}(\cdot | q)) }$$
Substituting this back into the bound yields: 
$$| \mathbb{E}_{o \sim \pi_\theta(\cdot | q^+)}[\mathcal{R}(o)] - \mathbb{E}_{o \sim \pi_\theta^{\mathbf{v}}(\cdot | q)}[\mathcal{R}(o)] | \le \sqrt{2 D_{\text{KL}}(\pi_\theta(\cdot | q^+) \parallel \pi_\theta^{\mathbf{v}}(\cdot | q))}$$
\end{proof}
Recall and simplify our DPO-style $\mathcal{L}(\mathbf{v}; q)$: 
$$\mathcal{L}(\mathbf{v}; q) = - \left[ \frac{1}{\hat{G}} \sum_{i=1}^{\hat{G}} \text{log}\text{ }\sigma \left( \beta\text{ }\text{log} \frac{\pi_\theta^{\mathbf{v}}(o_i^+|q)}{\pi_\theta(o_i^+|q)}-\beta\text{ }\text{log} \frac{\pi_\theta^{\mathbf{v}}(o_i|q)}{\pi_\theta(o_i|q)} \right) \right]. $$
\textbf{Connecting $\mathcal{L}(\mathbf{v}; q)$ to To the Bound.} $\mathcal{L}(\mathbf{v}; q)$ maps directly onto DPO \citep{rafailov2023direct} by establishing preference between the \emph{hinted} responses $o^+ \sim \pi_\theta(\cdot | q^+)$ relative to the \emph{plain} responses $o \sim \pi_\theta(\cdot | q)$, evaluated under $q$ with the reference model $\pi_\theta$. Within DPO, the implicit reward function parameterized by the model with activation steering $\pi_\theta^{\mathbf{v}}$ relative to the reference model $\pi_\theta$ is defined as: 
\begin{equation*}
r^{\mathbf{v}}(q, o) = \beta \log \frac{\pi_\theta^{\mathbf{v}}(o|q)}{\pi_\theta(o|q)}. 
\end{equation*}
Importantly, minimizing $\mathcal{L}(\mathbf{v}; q)$ performs maximum likelihood estimation of this implicit reward under the Bradley-Terry preference framework. As demonstrated by \citet{rafailov2023direct}, this estimation procedure aligns $\pi_\theta^{\mathbf{v}}(\cdot | q) \propto \pi_\theta(\cdot | q) \exp\left( \frac{1}{\beta} r^*(q, \cdot) \right)$ with the target (optimal) $\pi^*(\cdot | q)$ – here, defined as $\pi_\theta(\cdot | q^+)$ – that is, it pushes $\pi_\theta^{\mathbf{v}}(\cdot | q)$ to track $\pi_\theta(\cdot | q^+)$. Therefore, minimizing $\mathcal{L}(\mathbf{v}; q)$ is a surrogate for minimizing $D_{\text{KL}}(\pi_\theta(\cdot | q^+) \parallel \pi_\theta^{\mathbf{v}}(\cdot | q))$.

\section{Training Data and Hyperparameters}
\label{app-hyperparameters}
Due to limited computational resources, we follow prior work \citep{liao2026self, xia2026} to curate a condensed subset of the OpenR1-MATH-220k dataset \citep{openr1} as follows. Starting from the default 94k subset, we apply Math-Verify to filter out questions whose reference solution cannot be validly parsed, yielding a 64k subset. From this, 15k questions whose reference solution is under 4096 tokens are randomly selected and utilized as training data.

We implement ASH using \texttt{verl} \citep{sheng2025}, and for baselines, we adopt their publicly available implementations. Key training hyperparameters for reproducibility are listed in Table \ref{tab-4}. We run all experiments on 2 NVIDIA A100 80GB GPUs, and report the corresponding training times in Table \ref{tab-5}. A max prompt length of 1088 ensures that no questions are filtered. Empirically, our training logs indicate that the response truncation rate was less than \underline{\textcolor{red}{5\%}} when capping the maximum response length at 4096 tokens. This indicates that a max response length of 4096 does not affect our observations and conclusions while reducing computational expense. Following prior research \citep{zou2023representation}, we choose the \textbf{middlemost} layer of each model for activation steering, avoiding the extensive layer-wise sweeps that would otherwise demand high computational expense. We fix the scaling factor $\beta$ to 0.1, as in the original DPO.

\begin{table}[!ht]
\centering\footnotesize
\caption{Key Training Hyperparameters.}
\label{tab-4}
\begin{tabular}{lr}
\midrule
max prompt length           & 1088  \\
max response length         & 4096  \\
temperature                 & 1.0   \\
step size                   & 1e-6  \\
warmup steps ratio          & 0.1   \\
warmup style                & cosine\\
KL loss coefficient         & 0.0   \\
clip ratio (lower)          & 0.2   \\
clip ratio (upper)          & 0.2   \\
batch size                  & 128   \\
mini batch size             & 64    \\
micro batch size per GPU    & 8     \\
total epochs                & 1     \\
\midrule
\end{tabular}
\end{table}
\begin{table}[!ht]
\centering\footnotesize
\caption{Training Times of ASH and Baselines.}
\label{tab-5}
\begin{tabular}{lrlr}
\midrule
Model &Method & &Training Times \\
\midrule
\multirow[t]{5}{*}{Qwen3-1.7B-Base} &GRPO &-& $\sim$10 hours \\
& \multirow[t]{4}{*}{HA-GRPO}       &QUESTA & $\sim$12 hours \\
& &SAGE                                     & $\sim$18 hours \\
& &HILL                                     & $\sim$20 hours \\
& &ASH                                      & $\sim$15 hours \\
\midrule
\multirow[t]{5}{*}{Qwen3-4B-Base}   &GRPO &-& $\sim$20 hours \\
& \multirow[t]{4}{*}{HA-GRPO}       &QUESTA & $\sim$24 hours \\
& &SAGE                                     & $\sim$36 hours \\
& &HILL                                     & $\sim$40 hours \\
& &ASH                                      & $\sim$30 hours \\
\midrule
\end{tabular}
\end{table}

\section{Training Dynamics}
\label{app-logs}
In this section, we report supplementary training dynamics that offer deeper insight into the training patterns of ASH and the baselines, and further support the claims made in the paper.

\begin{figure}[!t]
\centering\includegraphics[width=\linewidth]{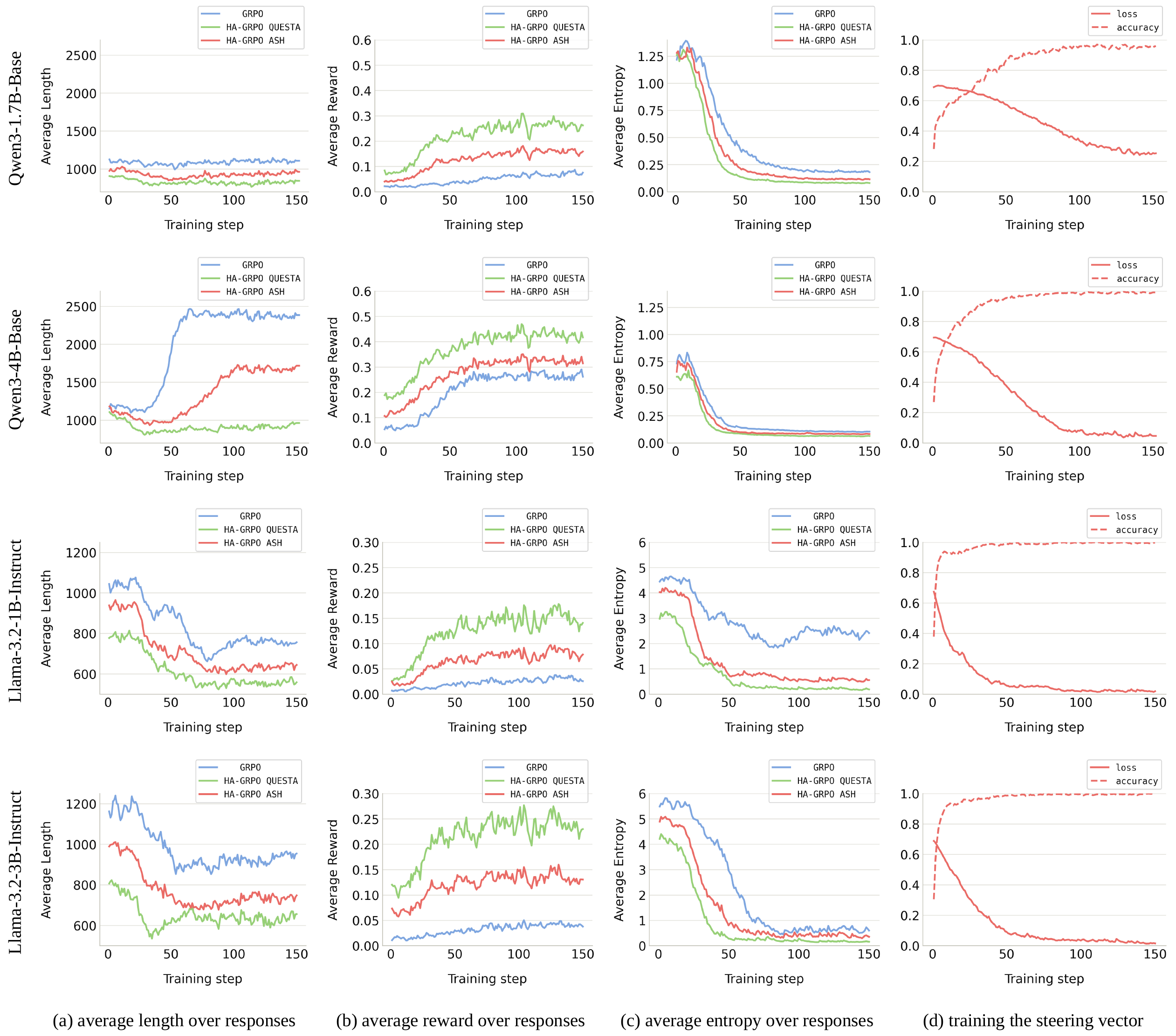}
\caption{Empirical training dynamics.}
\label{fig-logs}
\end{figure}

Figures \ref{fig-logs}a, b, and c display, respectively, the average response length, reward, and entropy of the generation probabilities per batch over the course of training. These metrics saturate during the final third of training, indicating that models have converged. Moreover, across three metrics, ASH consistently falls \textit{between} GRPO and QUESTA, reflecting the influence of the regularization introduced by joint training on both hinted questions and plain questions. Notably, the average response reward remains below 0.5 for Qwen3-1.7B-Base and Qwen3-4B-Base and 0.3 for Llama-3.2-1B-Instruct and Llama-3.2-3B-Instruct, highlighting the difficulty of questions in the training data.

Meanwhile, Figure \ref{fig-logs}d shows the loss $\mathcal{L}(\mathbf{v}; q)$ (Equation \ref{eqn-4}) and accuracy for training the steering vector, which decrease and increase as expected, respectively, indicating that the steering vector is trained properly, where accuracy is defined as the frequency of positive margins.

\section{On the pass@\texorpdfstring{$k$}{k} Metric}
\label{app-metric}
Here, we provide a thorough discussion of the pass@$k$ metric used in the paper.

\textbf{Definition.} pass@$k$ \citep{chen2021} measures the probability that at least one of $k$ independently generated responses from the model succeeds (i.e., produces a correct final answer) for a given question. Formally, given the question and $n$ responses to it, of which $c$ succeed, an unbiased estimator of pass@$k$ is:
$$\text{pass@}k = \mathbb{E}_{\text{questions}}\left[ 1 - \frac{\binom{n-c}{k}}{\binom{n}{k}} \right].$$
When $k = 1$, the metric reduces to the average success rate under temperature-controlled generation; when $k > 1$, it captures the model's ability to produce a correct final answer at least once across multiple attempts. We compute the metric with $n = 8$, using a temperature of 0.6 and a top-p of 0.95, and a fixed random seed of 0.

A model might fail to solve a difficult question on a single attempt yet still possess the latent ability to solve it given more tries; conventional metrics miss this distinction entirely. Meanwhile, pass@$k$ with $k > 1$ addresses this by considering $k$ generated responses and counting the question as solved if any one of those responses is correct, directly testing whether the correct reasoning path exists anywhere within the model's output distribution rather than just whether it's the most likely output.

Therefore, we use pass@$k$ to measure the expansion of a model's reasoning capability, and, to balance evaluation coverage against computational cost, we limit our evaluation to $k = 8$.

\section{Full pass@\texorpdfstring{$k$}{k}}
\label{app-curves}
\begin{figure}[!t]
\centering\includegraphics[width=\linewidth]{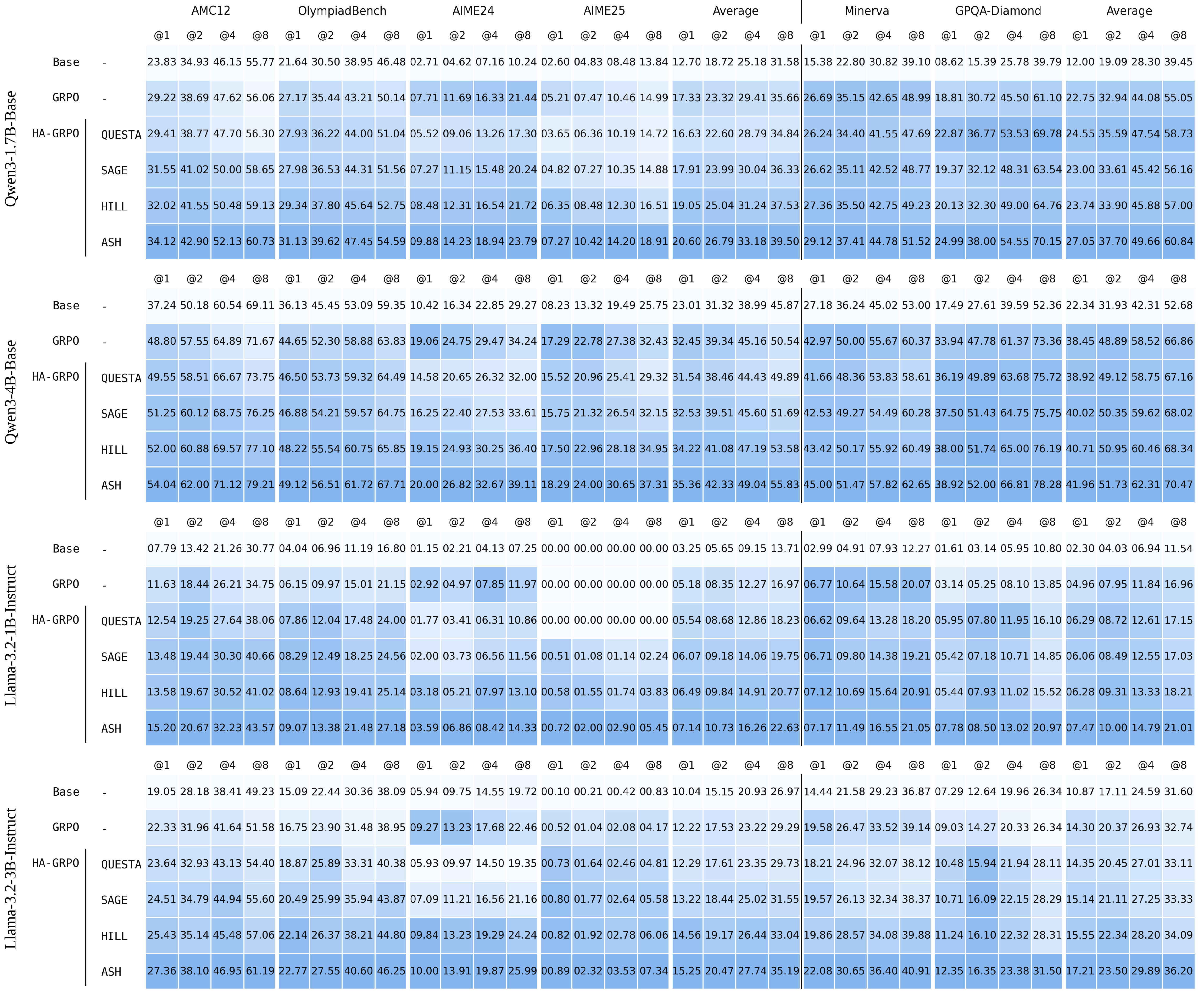}
\caption{pass@$k$ of models on six testing benchmarks. Shading ranks results within each column.}
\label{fig-full}
\end{figure}

We report full pass@$k$ (pass@1/pass@2/pass@4/pass@8) in Figure \ref{fig-full}, complementing the pass@8 alone presented in Table \ref{tab-1}. Remarkably, observations are consistent across all $k$.

\section{Failure Case Study}
\label{case}

To intuitively understand ASH's drawbacks, we explore cases where it fails to rectify model behavior. Table \ref{tab-6} shows a representative case on Qwen3-1.7B-Base and Qwen3-4B-Base.

We present a representative case that illustrates a failure mode of ASH, in which activation steering causes Qwen3-1.7B-Base and Qwen3-4B-Base to drift slightly from the input question (highlighted in \textcolor{red}{red}), leading to an incorrect reasoning path and final answer. We hypothesize that this phenomenon stems from the limited capacity of a single steering vector. Enriching this capacity with task- or question-specific steering vectors is a promising direction for future work.

\begin{table}[!ht]
\centering\scriptsize
\renewcommand{\arraystretch}{1.095}
\caption{A representative case that illustrates a failure mode of ASH.}
\label{tab-6}
\begin{tabular}{l|p{0.8\linewidth}}
\midrule
Model &
Question: $N$ is an integer, its representation in base $b$ is 777. Find the smallest positive integer $b$ such that $N$ is a fourth power of an integer in decimal notation. Ground Truth: 18\\
\midrule
Qwen3-1.7B-Base &
To solve this problem, we need to find the smallest positive integer $b$ such that the number $N$ represented in base $b$ as $777$ is a perfect power of an integer in decimal notation.
\newline Step 1: Convert $777$ from base $b$ to decimal notation
$$N = 7b^2 + 7b + 7 = 7(b^2 + b + 1)$$
Step 2: Set $N$ equal to a perfect power
\newline \textcolor{red}{We need $N$ to be a perfect square, so let $N = k^2$ for some integer $k$:}
$$7(b^2 + b + 1) = k^2$$
Step 3: Divisibility by 7
\newline Since 7 is prime and $7 \mid k^2$, we have $7 \mid k$. Let $k = 7m$. Then:
$$7(b^2 + b + 1) = 49m^2$$
$$b^2 + b + 1 = 7m^2$$
Step 4: Test small values of $m$ ($m = 1$):
$$b^2 + b + 1 = 7$$
$$b^2 + b - 6 = 0$$
$$b = \frac{-1 \pm \sqrt{1 + 24}}{2} = \frac{-1 \pm 5}{2}$$
$$b = 2 \quad \text{(since } b \text{ must be positive)}$$
\textcolor{red}{Since $b = 2$ is a positive integer, it is the smallest possible base.}
\newline Step 5: Verify the solution
$$N = 7 \cdot 2^2 + 7 \cdot 2 + 7 = 28 + 14 + 7 = 49 = 7^2$$
\textcolor{red}{Since $49 = 7^2$ is a perfect square, $b = 2$ satisfies the condition.}
\newline Final Answer: \boxed{2}
\newline \textit{Why it's wrong}: The problem asks for a fourth power, not a square. Base 2 also cannot contain the digit 7.
\\
\midrule
Qwen3-4B-Base &
To solve this problem, we need to follow these steps:
\newline Step 1: Convert the base $b$ representation to decimal
$$N = 7b^2 + 7b + 7$$
Step 2: Set up the fourth-power condition
$$7b^2 + 7b + 7 = k^4$$
Dividing both sides by 7:
$$b^2 + b + 1 = \frac{k^4}{7}$$
Since $b^2 + b + 1$ is an integer, $7 \mid k^4$, so $7 \mid k$. The smallest such $k$ is $k = 7$:
$$b^2 + b + 1 = \frac{2401}{7} = 343$$
$$b^2 + b - 342 = 0$$
Step 3: Solve the quadratic equation
$$b = \frac{-1 \pm \sqrt{1 + 1368}}{2} = \frac{-1 \pm \sqrt{1369}}{2} = \frac{-1 \pm 37}{2}$$
$$b = 18 \quad \text{or} \quad b = -19$$
We discard the negative root, so $b = 18$, which is greater than 7 and therefore a valid base.
Step 4: Compute $N$
$$N = 7(18)^2 + 7(18) + 7 = 2268 + 126 + 7 = 2401 = 7^4$$
\textcolor{red}{Since $k = 7$ is the smallest possible value of $k$, $N = 2401$ is the smallest fourth power of the form $777_b$. Therefore, the answer to the problem is $N$.}
\newline Final Answer: \boxed{2401}
\newline \textit{Why it's wrong}: The question asks for the smallest base $b$, not the value of $N$. The model finds $b = 18$ along the way but reports $N = 2401$ instead.
\\
\midrule
\end{tabular}
\end{table}

\end{document}